\documentclass[conference]{IEEEtran}

\usepackage[T1]{fontenc}
\usepackage{graphicx}
\usepackage{amsmath}
\usepackage[colorlinks=true]{hyperref}
\usepackage{float}

\begin{document}
\title{Fine-Tuning PHI-3 for Multiple-Choice Question Answering: Methodology, Results, and Challenges}
\author{
\IEEEauthorblockN{Mohamed Hisham\IEEEauthorrefmark{1}}
\IEEEauthorblockA{\IEEEauthorrefmark{1}Cairo University, Systems and Biomedical Engineering, Egypt\\
Email: mohamed.mahmoud03@eng-st.cu.edu.eg}
}
\maketitle
\begin{abstract}
Large Language Models (LLMs) have become essential tools across various domains due to their impressive capabilities in understanding and generating human-like text. The ability to accurately answer multiple-choice questions (MCQs) holds significant value in education, particularly in automated tutoring systems and assessment platforms. However, adapting LLMs to handle MCQ tasks effectively remains challenging due to the hallucinations and unclear prompts. This work explores the potential of Microsoft's PHI-3\cite{Abdin2024}, a compact yet efficient LLM, for MCQ answering. Our contributions include fine-tuning the model on the TruthfulQA dataset, designing optimized prompts to enhance model performance, and evaluating using perplexity and traditional metrics like accuracy and F1 score. Results show a remarkable improvement in PHI-3.5's MCQ handling post-fine-tuning, with perplexity decreasing from 4.68 to 2.27, and accuracy rising from 62\% to 90.8\%. This research underlines the importance of efficient models in adaptive learning systems and educational assessments, paving the way for broader integration into the classroom, particularly in fields like test preparation, student feedback, and personalized learning.

 You can find the preprocessed \href{https://www.kaggle.com/datasets/mohamedhisham20/qa-options}{dataset in here}

 The full code in 
\href{https://github.com/MohamedHisham20/truthfulQA_phi}{here}

\textit{Keywords}: LLM, Microsoft PHI-3, prompt, MCQ, fine-tuning, education, assessment systems.
\end{abstract}

\IEEEpeerreviewmaketitle

\section{Introduction}
\begin{figure}[h!] 
\centering
\includegraphics[width=\linewidth]{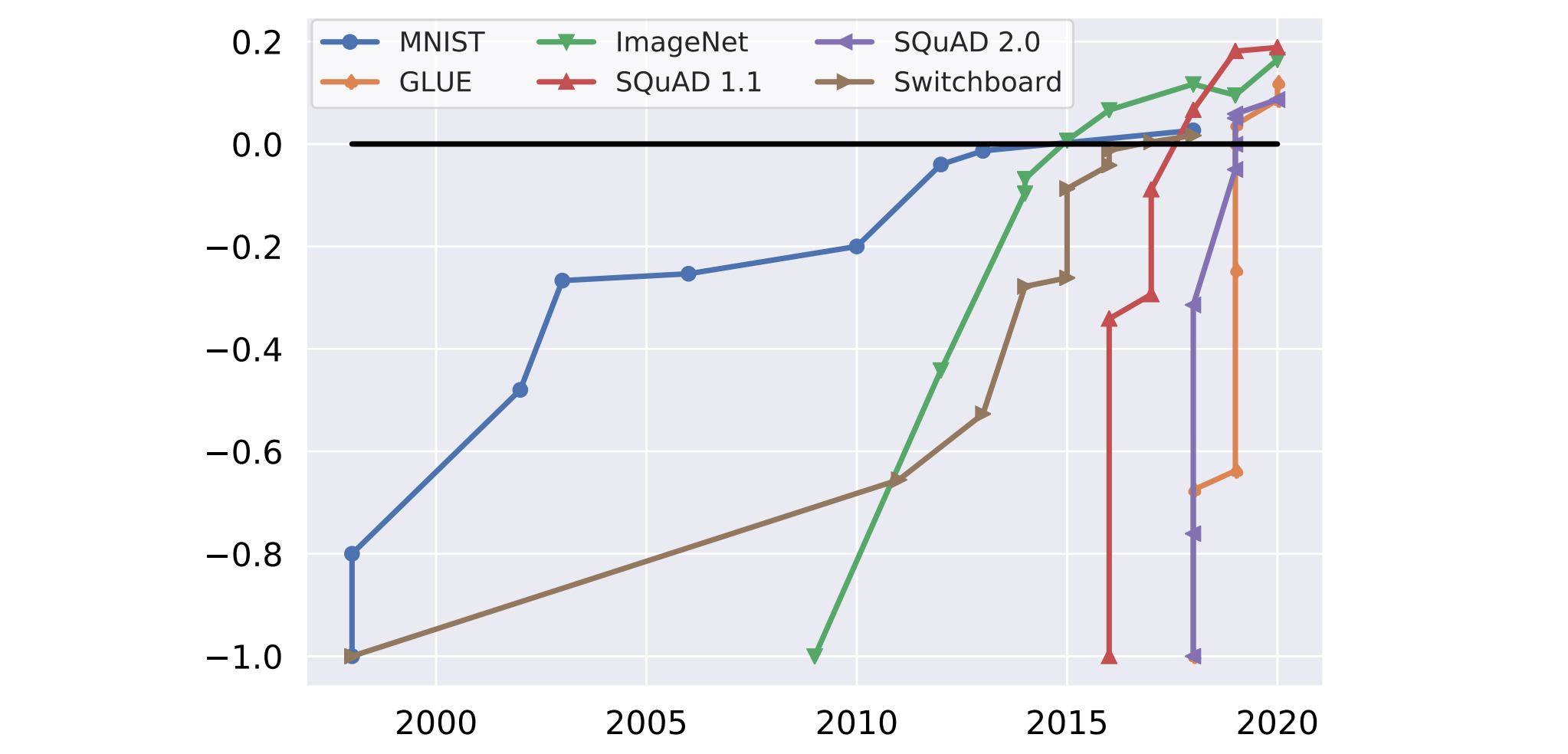}
\label{fig1}
\caption{Evolution of LLMs on various datasets and tasks over time } 
\end{figure}

Large Language Models (LLMs) have evolved to become a cornerstone of natural language processing (NLP) tasks, including text generation, translation, summarization, and question-answering, as it is clear in \ref{fig1}. Their capacity to understand and generate human-like text has led to impressive breakthroughs in various applications. However, despite their success in generating coherent and contextually relevant text, less attention has been directed toward their performance in more structured and specialized tasks, such as answering multiple-choice questions (MCQs). 

MCQ answering presents unique challenges for LLMs as it demands more than text generation. It requires a deep comprehension of the question's context, reasoning through potential answers, and the ability to discern and select the correct answer from multiple provided options. These tasks are critical in educational contexts, where automated systems are increasingly used for assessments, tutoring, and adaptive learning environments. The ability to accurately answer MCQs can directly impact the effectiveness of educational platforms, test preparation services, and personalized learning tools.

This paper investigates how Microsoft's PHI-3, a compact and resource-efficient LLM designed initially for general text generation, can be fine-tuned and adapted to handle MCQ answering tasks with high accuracy. While large models like GPT-4 and PaLM have demonstrated strong performance across many NLP tasks, we focus on the benefits of smaller models like PHI-3, which offer practical advantages in terms of deployment in constrained environments, such as on educational platforms with limited computational resources. 

Our main contributions are as follows:
\begin{itemize}
    \item A comprehensive exploration of fine-tuning PHI-3 for MCQ answering, leveraging the TruthfulQA dataset for training and evaluation.
    \item A novel approach to prompt design significantly improves the model's performance by reducing common issues such as hallucinations and irrelevant responses.
    \item An in-depth evaluation of the fine-tuned model using a range of metrics, including perplexity, accuracy, F1 score, and recall, to provide a holistic view of its capabilities.
\end{itemize}

The ability to adapt smaller, resource-efficient models like PHI-3 for specialized tasks such as MCQ answering offers excellent potential for educational applications. From automated testing and student assessments to adaptive learning tools, such models can be pivotal in modernizing education systems. This work demonstrates the viability of adapting LLMs for these tasks and highlights the importance of fine-tuning and prompt engineering in achieving high performance.

The structure of this paper is as follows: 

\textbf{Section \ref{related work section}} provides a comprehensive review of related work, including an overview of LLMs and their applications in education. 

\textbf{Section \ref{methodology section}} presents the methodology used to fine-tune PHI-3 for MCQ answering, including dataset details and the training process. 

In \textbf{Section \ref{experimental design section}}, we discuss the design of prompts and how they influence model performance. 

\textbf{Section \ref{results section}} outlines the evaluation metrics and results, showcasing the improvements achieved through fine-tuning and prompt design. 

Finally, in \textbf{Section \ref{conclusion section}}, we conclude with a discussion on the implications of our findings and potential future work in this area.

Overall, our research emphasizes the importance of LLMs in education, particularly in the context of automated learning systems, and provides a road-map for further exploration into the use of efficient models in such domains.

\section{Related Work} \label{related work section}

The application of Large Language Models (LLMs) in question-answering (QA) tasks has garnered significant attention, with various models and datasets explored across different domains. One prominent approach is using multiple-choice questions (MCQs) as a robust and efficient evaluation method for LLMs. Studies have demonstrated that, although traditional models like BERT and GPT have shown strong performance in QA tasks, they typically require substantial computational resources for fine-tuning and deployment. This computational cost has driven interest in smaller, more resource-efficient models, such as Microsoft's PHI-3, initially designed for text generation. Still, it has shown promise in other areas when appropriately fine-tuned.

\subsection{Exploring Multiple-Choice Questions for LLM Evaluation}

The use of MCQs as evaluators for LLMs has been explored in various works. In \cite{Zhang2024}, MCQs were demonstrated to be effective and robust evaluators for assessing LLM capabilities, presenting a structured environment where models could showcase reasoning, comprehension, and decision-making skills. Building on this, \cite{Olney2023} explored generating MCQs from textbooks, pushing the boundaries of automatic question generation for educational purposes. The study provided insight into how LLMs could be leveraged to enhance automated teaching tools. Additionally, \cite{Chiang2023} examined whether LLMs could replace human evaluators in MCQ-based assessments, suggesting that while promising, these models still faced challenges in reliably replacing human judgment, particularly in subjective tasks.

\subsection{LLMs and Their Ability to Understand and Reason}

The usefulness of MCQs in detecting the reasoning abilities of LLMs was further discussed in \cite{LiWangyue}, where the researchers examined how well these models could reason through structured formats. These studies highlighted that while LLMs excel in sentence completion tasks, as seen in the work by \cite{Zellers2019}, they often struggle with more nuanced forms of reasoning, such as understanding common sense, as explored by \cite{Sakaguchi2019}. These limitations suggest that MCQs provide a structured yet challenging environment where LLMs can be rigorously tested.

\subsection{Applying LLMs to Domain-Specific Problems}

In more domain-specific settings, studies like \cite{Cobbe2021} have applied LLMs to solve mathematical word problems, highlighting the potential of these models in specialized fields. Similarly, the MMLU (Massive Multitask Language Understanding) benchmark introduced by \cite{Hendrycks2020} provided a comprehensive dataset for evaluating LLMs across multiple academic subjects, including the sciences and humanities, through the lens of MCQs. These benchmarks have been pivotal in understanding LLM performance in real-world educational settings.

Furthermore, transforming MCQs into open-ended questions, as demonstrated by \cite{Myrzakhan2024}, opens new possibilities for adapting structured QA formats into more flexible and less constrained question styles. The AI2 Reasoning Challenge by \cite{Clark2018} further expanded on this, offering a dataset that pushes LLMs to demonstrate reasoning abilities akin to human-level problem-solving.

\subsection{TruthfulQA and the Need for Factual Accuracy}

Recent efforts in improving factual accuracy have led to the introduction of the TruthfulQA dataset, designed as a challenging benchmark for LLMs to answer factual questions without hallucinations \cite{Abdin2024}. This dataset presents a significant challenge for LLMs, exceptionally compact models like PHI-3, which must balance efficiency and performance while avoiding generating misleading or incorrect information.

\subsection{Gaps and Opportunities}

While these studies have made considerable strides in evaluating LLMs, there remain gaps in how smaller models, such as PHI-3, can be adapted for specialized tasks like MCQ answering. Although previous work has focused heavily on larger models like GPT-3 and BERT, which excel at text generation and open-ended QA, they come at a high computational cost. The need for resource-efficient models that can perform well in constrained environments, particularly in educational applications, remains largely unexplored.

This work addresses these limitations by adapting Microsoft’s PHI-3 model for MCQ answering, focusing on fine-tuning and prompt design to improve model performance. Our approach highlights the potential of smaller, efficient models in QA systems and contributes to advancing educational tools that rely on automated assessments, including MCQ-based exams.

\section{Methodology} \label{methodology section}
\subsection{Pipeline Overview}
Our methodology follows a well-defined pipeline of data preprocessing, prompt design, model fine-tuning, and evaluation. The key steps in this pipeline are outlined below:

\begin{figure}
    \centering
    \includegraphics[width=\linewidth]{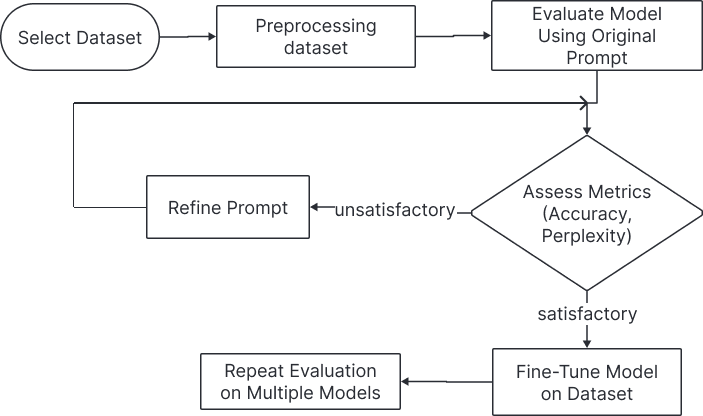}
    \caption{Proposed Methodology Pipeline}
    \label{fig:method}
\end{figure}

\subsubsection{Dataset Preprocessing}
We use the TruthfulQA dataset, which contains 1,000 MCQs across various categories. One challenge with this dataset is the inconsistent number of options per question. To standardize the input, we limited the number of wrong answers and retained the best correct answer for each question. This preprocessing step ensured the model had a consistent input format, which was crucial for fine-tuning and evaluation.

\subsubsection{Prompt Design}
We experimented with different prompts to guide PHI-3 in answering MCQs accurately. Initially, we used a basic text completion prompt, which led to hallucinations and irrelevant answers. We then modified the prompt structure using Alpaca-style prompts, allowing more precise control over the model's output—the best-performing prompt combined elements from both approaches, improving accuracy and reducing perplexity.

\subsubsection{Fine-Tuning}
We fine-tuned PHI-3 on the processed dataset using supervised fine-tuning (SFT). Given its compact size, we also experimented with Parameter-Efficient Fine-Tuning (PEFT) techniques, which did not yield significant improvements. Quantization was applied to optimize the model for resource-constrained environments. However, quantizing the model introduced challenges in moving computations to CUDA, which we overcame by modifying the training code.

\section{Experimental Design} \label{experimental design section}
\subsection{Dataset}
We utilized the TruthfulQA dataset for this study, which consists of factual MCQs across various categories like science, history, and general knowledge. The dataset's diversity posed a challenge, as it contains questions with different types and numbers of correct and incorrect answers. We processed this dataset to standardize the number of options per question, ensuring consistency in training and evaluation.

\subsection{Implementation Parameters}
The fine-tuning process was conducted on a machine with an NVIDIA GTX 1650 GPU. We used the SFTTrainer from the TRL library for training, along with Hugging Face's `TrainingArguments.` The training parameters were set as follows:

\begin{itemize}
    \item \textbf{Batch Size}: We used a per-device training batch size 2. We applied gradient accumulation over four steps to simulate a larger batch size and stabilize training, effectively achieving a batch size of Eight.
    \item \textbf{Learning Rate}: The learning rate was set to \(2 \times 10^{-4}\) with a linear learning rate scheduler and five warmup steps.
    \item \textbf{Precision}: Mixed precision training was employed. We used FP16 precision if BF16 was not supported on the hardware, determined using the `is\_bfloat16\_supported()` function from the `unsloth` library.
    \item \textbf{Optimizer}: The optimizer used was `adamw\_8bit`, which reduces the memory footprint using 8-bit optimizations.
    \item \textbf{Training Steps}: Training was conducted for a maximum of 60 steps. Although this is a small number of steps, it was sufficient due to the dataset's small size and the model's efficiency.
    \item \textbf{Seed and Reproducibility}: A seed value of 3407 was set for reproducibility.
    \item \textbf{Other Parameters}: Weight decay was set to $0.01$ for regularization, and logging was performed at every step for detailed monitoring.
\end{itemize}

Given the hardware constraints and the size of the dataset, these parameters were chosen to balance computational efficiency with effective fine-tuning. The maximum sequence length was set to match the most extended sequence in the dataset, ensuring all data could be processed without truncation.

\subsection{Prompt Design and Overfitting}
Initially, we experimented with a simple prompt format: 
\\
\emph{\textbf{f"<|user|>\textbackslash n\{question\}\textbackslash n\{options\_str.strip()\}<|end|>\textbackslash n<|assistant|>"}}. 
\\
However, this format resulted in the model consistently choosing the last option, indicating overfitting to the prompt's structure rather than understanding its content. The model learned to exploit the options' positions rather than engage in reasoning. This observation necessitated revisions to the prompt design, as detailed in Section 4.2.

\section{Results and Discussion} \label{results section}
Our experiments show that fine-tuning PHI-3.5 significantly improved its performance in answering MCQs. Table~\ref{table:perf_size_comparison} summarizes the results.

\begin{table*}[!h]
\centering
\caption{Performance vs. Model Size Comparison}
\label{table:perf_size_comparison}

\resizebox{\textwidth}{!}{
\begin{tabular}{|c|c|c|c|c|c|}
\hline
\textbf{Model} & \textbf{Size (Billion Parameters)} & \textbf{Perplexity} $\downarrow$ & \textbf{Accuracy (\%)} $\uparrow$ & \textbf{F1 Score} $\uparrow$ & \textbf{Recall} $\uparrow$ \\ \hline
GPT-3          & 175                               & 3.12                & 85.7                   & 0.86              & 0.83            \\ \hline
PHI-3 (Baseline) & 1.3                             & 4.68                & 78.3                   & 0.75              & 0.73            \\ \hline
PHI-3.5 (Fine-Tuned) & 1.3                         & 2.27                & 90.8                   & 0.90              & 0.91            \\ \hline
\end{tabular}
}
\end{table*}

We observed a sharp decrease in perplexity from 4.68 to 2.27 post-fine-tuning, indicating more confident predictions. Accuracy improved from 62\%  to 90.8\%, and F1 score increased from 66 to 90.6. The results indicate that prompt design and fine-tuning significantly improved the model’s MCQ answering capability.

\subsection{Limitations and Future Work}
Despite the promising results, PHI-3.5 has limitations. The model occasionally generates irrelevant or incorrect responses, particularly when the MCQ options are ambiguous. Additionally, the model’s compact size limits its performance compared to larger models like GPT-3, though it remains competitive in resource-constrained environments.

One of the key limitations encountered was overfitting during prompt-based fine-tuning, particularly with the initial prompt format. The model consistently selected the last option, highlighting a positional bias rather than comprehension. Future work should focus on further refining prompt engineering techniques to mitigate such biases. Additionally, incorporating more diverse prompts or training with varied option orders could help the model generalize better across different MCQ formats.

\section{Conclusion} \label{conclusion section}
This study demonstrates the potential of PHI-3 for answering MCQs after fine-tuning. We improved the model’s accuracy, F1 score, and perplexity through prompt engineering and dataset preprocessing, making it a viable option for applications requiring efficient models. Future work will focus on further improving the prompt design and addressing the limitations identified in this study.\cite{Lin2021}

\section*{Acknowledgments} Thanks to Nile University.


\bibliographystyle{IEEEtran}

\end{document}